\documentclass[10pt,twocolumn,letterpaper]{article}

\usepackage[pagenumbers]{cvpr} % To force page numbers, e.g. for an arXiv version

\usepackage{graphicx}
\usepackage{amsmath}
\usepackage{amssymb}
\usepackage{booktabs}
\usepackage{multirow}

\usepackage[pagebackref,breaklinks,colorlinks]{hyperref}

\usepackage[capitalize]{cleveref}
\crefname{section}{Sec.}{Secs.}
\Crefname{section}{Section}{Sections}
\Crefname{table}{Table}{Tables}
\crefname{table}{Tab.}{Tabs.}

\def\cvprPaperID{JoyBoy Team} % *** Enter the CVPR Paper ID here
\def\confName{CVPR}
\def\confYear{2022}

\begin{document}

\title{InternVideo-Ego4D: A Pack of Champion Solutions to Ego4D Challenges}

\author{
Guo Chen$^{1,2\dagger}$, 
Sen Xing$^{1,3\dagger}$,
Zhe Chen$^{1,2\dagger}$,
Yi Wang$^{1\dagger}$, 
Kunchang Li$^{1,4}$, 
Yizhuo Li$^{1,5}$, 
Yi Liu$^{1,4}$ \\
Jiahao Wang$^{2}$,
Yin-Dong Zheng$^{2}$,
Bingkun Huang$^{1,2}$,
Zhiyu Zhao$^{1,2}$,
Junting Pan$^{1,6}$ \\
Yifei Huang$^{1}$,
Zun Wang$^{1,7}$,
Jiashuo Yu$^{1}$,
Yinan He$^{1}$,
Hongjie Zhang$^{1}$ \\
Tong Lu$^{2*}$,
Yali Wang$^{4,1*}$,
Limin Wang$^{2,1*}$,
Yu Qiao$^{1*}$ \\
% \vspace{2mm} 
\\
$^1$Shanghai AI Laboratory, $^2$Nanjing University, $^3$Tsinghua University \\
$^4$Shenzhen Institute of Advanced Technology, Chinese Academy of Sciences\\ 
$^5$The University of Hong Kong,
$^6$The Chinese University of Hong Kong \\
$^7$The Australia National University
\\
%{\tt\small qiaoyu@pjlab.org.cn} \\
}
\newcommand\blfootnote[1]{%
\begingroup
\renewcommand\thefootnote{}\footnote{#1}%
\addtocounter{footnote}{-1}%
\endgroup
}

\maketitle

% !TeX spellcheck = en_US
%!TEX root=../main.tex
\begin{abstract}
In this report, we present our champion solutions to five tracks at Ego4D challenge. We leverage our developed \textbf{InternVideo}, a video foundation model, for five Ego4D tasks, including Moment Queries, Natural Language Queries, Future Hand Prediction, State Change Object Detection, and Short-term Object Interaction Anticipation.
InternVideo-Ego4D is an effective paradigm to adapt the strong foundation model to the downstream ego-centric video understanding tasks with simple head designs.
In these five tasks, the performance of InternVideo-Ego4D comprehensively surpasses the baseline methods and the champions of CVPR2022, demonstrating the powerful representation ability of InternVideo as a video foundation model. Our code will be released at \url{https://github.com/OpenGVLab/ego4d-eccv2022-solutions}.

\end{abstract}

\blfootnote{$\dagger$ equal contribution. * corresponding author.}

% !TeX spellcheck = en_US
%!TEX root = ../main.tex
\section{Introduction}
\label{sec:introduction}

\begin{figure}[!t]
\centering
  \includegraphics[width=0.5\textwidth]{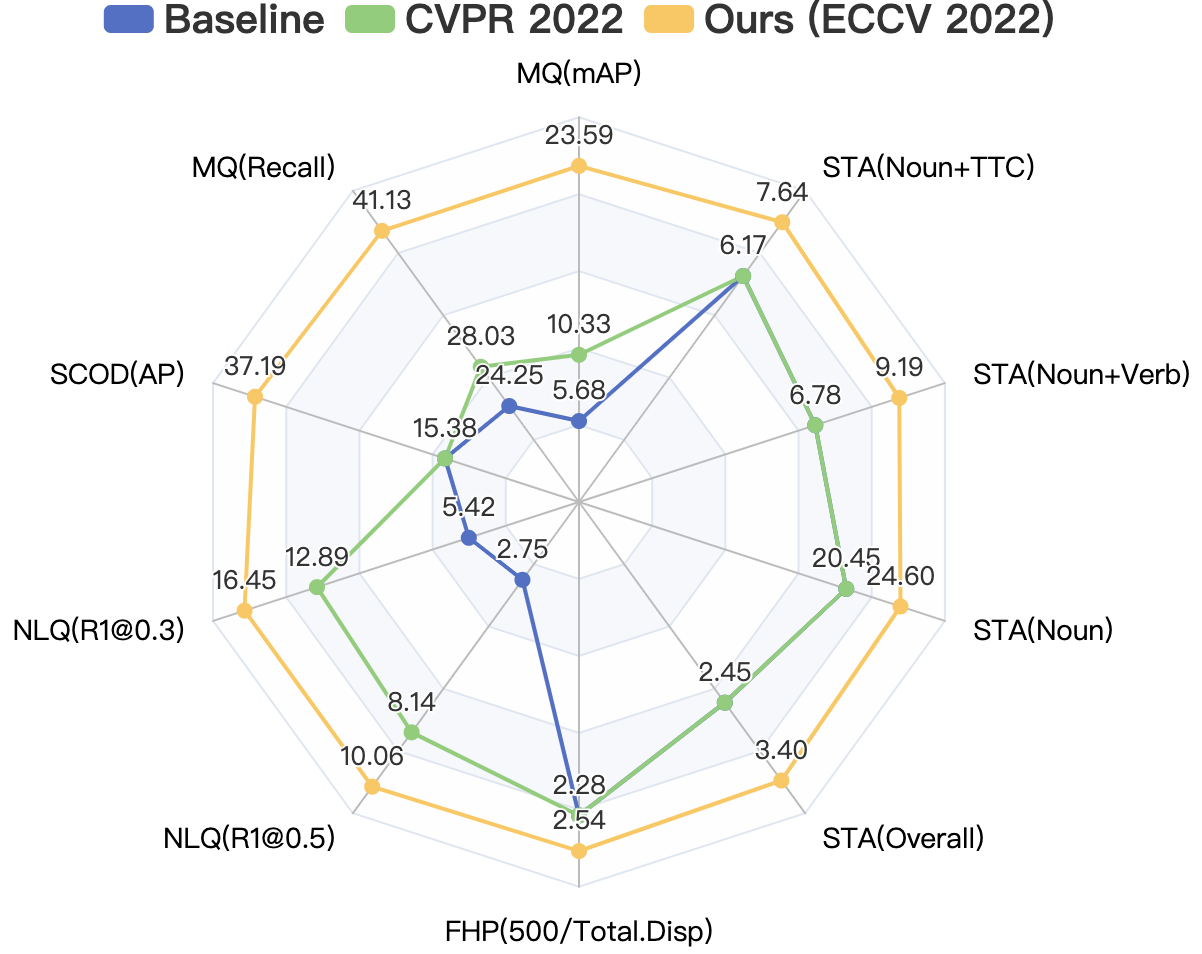}
  \caption{
For the 10 evaluation indicators of 5 tasks, InternVideo has achieved all-around improvement, demonstrating its astounding performance and strong support for downstream tasks.
}
  \label{fig:radar}
\end{figure}

% 1、introduce ego4d challenges %
% 2、introduce the performance in CVPR2022 %
% 3、we take part in ECCV2022 %
% 3、we take part in ECCV2022 %
Ego4D~\cite{ego4d} is the latest large-scale egocentric video understanding dataset presented by Facebook AI Research (FAIR). 
Different from the previous video understanding datasets, Ego4D unintentionally collects persistent 3D egocentric data, including the camera wearer's physical surroundings, interactive objects and actions, and high-level social behaviors. 
It aims to catalyze the next era of research in first-person visual perception. 
The accompanying 5 benchmark tasks open new research directions and stimulate research broadly from around the world to a large extent. 
These benchmark tasks cover the basic components of egocentric perception—indexing past experiences, analyzing current interactions, and predicting future activities. To promote Ego4D and explore its research methodologies, several challenges are organized.

In the Ego4D ECCV2022 Challenge, we joined in five tracks Moment Queries, Natural Language Queries, Future Hand Prediction, State Change Object Detection, and Short-term Object Interaction Anticipation. As shown in Table~\ref{fig:radar}, we won all championships in these tasks. We share our solutions in this technical report.

Despite efforts to adapt cutting-edge task heads to our participated tracks, we leverage our developed \textbf{InternVideo}, a video foundation model, to support our competitions. As it is an effective paradigm to exploit a strong foundation model to address downstream tasks, simplifying head designs. InternVideo involves both masked autoencoder and multimodal learning, and we chiefly employ its two components VideoMAE \cite{videomae} and UniFormer \cite{uniformer}. VideoMAE offers a spatio-temporal representation using a vision transformer encoder by a masked video reconstruction pretask, while UniFormer integrates local spatio-temporal modeling into transformers for efficient video representation learning. Besides of these two backbones, we also explore others, \eg\ Swin, EgoVLP, ResNet, and more, for comparisons or fusion.

In the remaining of this report, we will introduce the related work first. And then we will detail our solutions along with experiments to each joined Ego4D track, after briefing the employed VideoMAE and UniFormer. Finally, we discuss the limitations of our work and conclude this paper.
\section{Related Work}

% \subsection{Ego4D CVPR2022 Challenges }

\subsection{Backbone in Video Understanding}
% Currently, video understanding tasks mainly focus on extracting and understanding timing information and spatial information. Most video understanding tasks can be divided into upstream and downstream modules. The upstream part is mainly composed of a backbone, and the downstream part is a specific head designed for different tasks.

% In this way, the design of decoupling upstream and downstream is simple and feasible. With this idea, coupled with our self-supervised training method, backbones with powerful feature extraction capabilities and trained on many datasets can be directly applied. Among multiple downstream tasks, the effect of SOTA can be achieved on many downstream tasks only by finetuning on the corresponding dataset.

3D Convolutional Neural Networks (CNNs) have been dominant in video understanding \cite{c3d,i3d}.
Due to the difficult optimization problem and large computation of 3D convolution,
great efforts have been made to factorize 3D convolution.
R(2+1)~\cite{r(2+1)d}, P3D~\cite{p3d} and TANet~\cite{tanet} divide the 3D convolution into 1D temporal convolution and 2D spatial convolution,
while CSN~\cite{csn}, X3D~\cite{x3d} and MoViNet~\cite{movinet} propose 3D channel-separated convolution.
However,
3D convolution struggles to capture long-range dependencies because of the local receptive field~\cite{non_local}.
Inspire by the success of Vision Transformers (ViTs) in the image domain \cite{vit,pvt,swin},
researchers try to apply global attention for spatiotemporal modeling \cite{timesformer,video_transformer,vidtr,vivit,x_vit,motionformer}.
For efficient and effective video understanding,
MViT~\cite{mvit} introduces a hierarchical structure with pooling self-attention,
VideoSwin~\cite{video_swin} extends window self-attention to 3D space and UniFormer~\cite{uniformer} proposes to unify convolution and self-attention for better accuracy-computation trade-off.

\subsection{Temporal Action Localization}
\textbf{Temporal Action Localization (TAL)} is aimed at detecting the boundaries and categories of the action segments in untrimmed videos.
Current temporal action detection methods can be divided into one-stage and two-stage methods. 
The two-stage methods decouple the tasks of generating proposals and classifying actions. For instance, \cite{bsn,bmn,bsn++,dcan} use a flexible way called boundary matching to generate high-quality proposals. They predict each frame's start and end confidence, then match the frames with high start and end confidence to generate the proposals and evaluate their confidence.
The one-stage methods generate action proposals and corresponding action labels simultaneously in a single model. Recently, \cite{afsd,tadtr-e2e,baisctad} explores the end-to-end training methods and yields outstanding performance. However, considering the end-to-end training overhead for video data, utilizing a feature-based one-stage detection method is a more efficient and convenient option. Among these methods, VSGN \cite{vsgn} adopts GNN to aggregate multi-scale temporal features to generate action predictions based on dense anchors. ActionFormer \cite{actionformer} builds a transformer network to predict the offsets of the start and end position through a well-implemented anchor-free mechanism.
In the Ego4D challenges, the \textbf{Moment Queries (MQ)} task aims to query the specific moments consistent with the TAL task. We adopt VSGN~\cite{vsgn} and ActionFormer~\cite{actionformer} as our detection heads to validate the performance of VideoIntern on this task.

\subsection{Video Temporal Grounding}
\textbf{Video Temporal Grounding (VTG)}, which needs to retrieve video segments using natural language queries, was introduced in~\cite{natural_language_lisa,natural_language_gaojiyang}. Early works explored how to utilize text queries. One method is metric learning based, such as \cite{mmn}, which uses metric learning loss functions with the distance as the similarity measurement to match the given text queries to the right video moment. 
The other is detection-based, transforming the text query as a dynamic filter as a condition for extracting the temporal feature. It can inherit excellent mechanisms from temporal action detection and object detection. VSLNet~\cite{vslnet} adopts the context-query attention module to predict video segments corresponding to the text queries. 
In the Ego4D challenges, the \textbf{Natural Language Queries (NLQ)} task aims to localize the correct temporal segments through natural language queries, consistent with the VTG task. We use VSLNet~\cite{vslnet} as our grounding head to validate the performance of VideoIntern on this task.

\subsection{Spatio-Temporal Action Localization}
The purpose of \textbf{Spatio-Temporal Action Localization (STAL)} is to predict people's location in keyframes and classify the person's ongoing actions in videos. At present, most methods divide this task into two subtasks. They first generate human bounding boxes in keyframes and then extract 3D ROI features for these boxes to classify actions.
The generated people boxes by Faster R-CNN \cite{faster-rcnn} is widely used in current STAL methods. Considering that there may be various actors or objects in a keyframe, these methods, such as \cite{alphaction, acar}, focus on capturing the relations and contexts in different levels between the objects in the video.

ViT~\cite{vit} uses the global self-attention mechanism to mix the spatial patches. \cite{vivit,timesformer} replaced spatial self-attention blocks with Spatio-temporal self-attention blocks so that ViT can achieve cube-level dense spatiotemporal interaction. Due to the full spatiotemporal interaction implemented in the backbone, the well-designed interaction modules in the STAL methods become unnecessary.
% \textbf{[TODO: describe performance with VideoIntern on STAL]}. 
Compared with the STAL task, the \textbf{Short-term Object Interaction Anticipation (STA)} task in Ego4D challenges has a similar form but different predicting requirements. Given a pre-condition clip, it needs first to forecast the bounding boxes of the objects that will be interacted with, then forecast the noun category of objects, the verb category of the interaction, and the time to contact objects. We use a similar manner to Spatio-temporal action localization to complete this track.

\subsection{Object Detection}
\textbf{Object Detection} is a classic 2D computer vision task to predict the bounding box regression and pixel-level classification results. 
The traditional CNN-based object detectors have been widely studied in the past years, such as R-CNN~\cite{girshick2015region}, Fast R-CNN~\cite{girshick2015fast}, Faster R-CNN~\cite{faster-rcnn}, and so on.
In recent years, transformer networks have become popular, and detectors based on transformers~\cite{detr,vitadapter,dino} have emerged as the times require. 
DETR~\cite{detr} and deformable DETR~\cite{zhu2020deformabledetr} used the transformer decoder to perform end-to-end object detection.
AdaMixer~\cite{adamixer} proposed a fast-converging decoding module in query-based detector based on sparse sampling and dynamic MLP mixer.
DINO~\cite{dino} made a series of optimizations and achieved state-of-the-art performance on the COCO~\cite{cocodet} dataset.

% !TeX spellcheck = en_US
%!TEX root=../main.tex
% \clearpage
\section{Methodology}
% \subsection{Backbone}

% VideoMAE~\cite{videomae,facebook_videomae} is a typical video backbone that uses the masking strategy for self-supervised training. Unlike ImageMAE~\cite{ImageMAE}, the video information is redundant in time, which allows the model to use a higher mask rate and achieve good results. The ViT model is used in VideoMAE, which uses tube-masking on a series of input frames to prevent information leakage caused by timing information redundancy, and the mask rate is as high as 90\%.

% The biggest advantage of VideoMAE is its self-supervision feature. As we all know, the largest number of social media on the Internet today are pictures and videos. What we lack is never data but the lack of annotations of these massive data. Although videos are often accompanied by some explanatory text, it is unreasonable to treat them as data annotations, because these explanatory texts are very different in style and quality. Therefore, using a model with self-supervised properties for training has huge advantages. A large amount of data is conducive to the convergence of the Transformer network, on the other hand, there are usually some gaps in data from different sources. Self-supervised training on different data is beneficial to improve the generalization ability of the model.

We choose CSN\cite{csn}, VideoMAE~\cite{videomae}, and UniFormer~\cite{uniformer} as our backbones for feature extraction. 
These three backbones have different architecture designs: pure convolution network, pure transformer network, and convolution-transformer hybrid network. 
We hypothesize that the representations from these three backbones are different and complementary. 
We employ different task-specific heads to complement different tasks.

\begin{table}[t]
\centering
\small
\setlength\tabcolsep{0.4mm}
\begin{tabular}{c|ccc}
\toprule
Backbone & ir-CSN-152~\cite{csn} & VideoMAE-L~\cite{videomae} & UniFormer-B~\cite{uniformer} \\
\midrule 
Frames & 32 & 16 & 16 \\
Res & 224 & 224 & 320  \\ 
WD & 1e-4 & 0.05 & 0.05  \\ 
\bottomrule

\end{tabular}
\caption{The finetuning settings of ir-CSN-152, VideoMAE-L, and UniFormer-B. ``Res'' and ``WD'' are short for resolution and weight decay.}
\label{table:finetune-setting}
\end{table}

% \subsubsection{Results} \label{pretrain-perform}

% We train these models on the verb and noun subsets and use the Top-1 and Top-5 accuracy of the validation set to measure the quality of finetuning. The results are shown in Table~\ref{table:finetune-performance}. 

\begin{table}[t]
\centering
\small
\setlength\tabcolsep{3 mm}
\begin{tabular}{c|cc|cc}
\toprule
 \multirow{2}{*}{Method}& \multicolumn{2}{c|}{Verb}  & \multicolumn{2}{c}{Noun}  \\
 & Top-1 & Top-5 & Top-1 & Top-5 \\
\midrule 
ir-CSN-152~\cite{csn} & 43.25 & $-$ & $-$ & $-$ \\
VideoMAE-L~\cite{videomae}& 52.51 & 86.05 & 33.41 &85.51  \\ 
UniFormer-B~\cite{uniformer} & 49.30 & 83.61 & $-$ & $-$ \\ 
\bottomrule

\end{tabular}
\caption{The finetuning performance of ir-CSN-152, VideoMAE-L, and UniFormer-B on our verb and noun validation set.}
\label{table:finetune-performance}
\end{table}

\subsection{Pre-training}

The used backbones have been pre-trained on action recognition datasets already. Specifically, CSN~\cite{csn}, VideoMAE~\cite{videomae}, and UniFormer~\cite{uniformer} are pre-trained on IG65M~\cite{ig65m}, K700~\cite{kinetics}, and K600~\cite{kinetics}, respectively.

These backbones need further finetuning as we find there is a distribution gap between data used in the pre-trained datasets and egocentric ones. 
Using the pre-trained backbones directly would lead to performance degradation in the egocentric video. 
To alleviate the negative effects brought by this gap, we finetuned these backbones on the Ego4D~\cite{ego4d} training set. 
Specifically, we adopt the clip-level annotations of EgoVLP~\cite{egovlp} on the training set for rapid development. 

In training on the Ego4D dataset, we finetune two model variants for each backbone using two types of annotations, considering that Ego4D has both the verb and noun category annotations for short clips in the video. 
Intuitively, models trained by the verb and noun annotations can capture action and scene information, respectively. 
Taking the verb annotation as an example, we preserve all the videos that contain the verb annotations. 
For videos containing multiple verb categories, we treat them as single-category annotations for multiple videos (the same video). 
To construct rapidly a validation set to evaluate the performance of fine-tuning,
we sample 5\% of the videos of each class as the validation set. 
Finally, we use the standard single-class action recognition training method to finetune the backbones with our constructed verb and noun datasets.

\subsection{Experiments}

We use ir-CSN-152, VideoMAE-L, and UniFormer-B as our backbones. These models are trained for 10 epochs by AdamW optimizer and cosine schedule. During fine-tuning, we set the batch size to 256 and the learning rate to 5e-4.
The other fine-tuning setting is shown in Table~\ref{table:finetune-setting}.

We train these models on the verb and noun subsets and use the Top-1 and Top-5 accuracy of the validation set to measure the quality of finetuning. The results are shown in Table~\ref{table:finetune-performance}. Due to its high performance, we mainly adopt VideoMAE-L as the feature extractor for the two temporal localization tasks, NLQ and MQ.

\section{Track 1: Moment  Queries}
\subsection{Problem Definition}
The Moment Queries track serves as the Episodic Memory task to query the moments of some high-level activities or ``moment'' names (that can be transformed into discrete labels). 
For an untrimmed egocentric video, we denote it as $U=\{u_t\}^{l_{v}}_{t=1}$, where $l_v$ indicates the length of the video and  $u_t$ is the $t$-th frame. We denote the temporal annotation of action instances as $\Psi_{\rm g}=\{\varphi_n=(t_{\rm s},t_{\rm e},c)\}^{N_{\rm g}}_{n=1}$ in the video $S_v$ which has  $N_{\rm g}$ instance. $t_{\rm s}$, $t_{\rm e}$, and $c$ are the start, end boundary, and categories of the instance $\varphi$, respectively. The model generates predicted moment segments that should cover $\Psi_{\rm g}$ with high recall and high temporal overlapping.

\subsection{Approach}

Due to the length of each untrimmed video coverage ratio of the activities being large in the MQ dataset, we adopt the two-stage method to localize temporal segments. 
Specifically, we first finetune our backbone with MQ labels and then extract the pooled Spatio-temporal features as the input of localization heads.
For validating comprehensively the feature extracted by our backbones, we use multiple localization heads as our candidate task-specific heads to observe the experimental results.
% In the following, we will describe the used backbones first and then describe how we aggregate video features from different backbones for performance. After that, we give settings and implementations about how to adapt the employed task heads to these two episodic memory tasks.

\textbf{Fine-tuning Settings.} Since the gap between different datasets, there are multiple routes for transfer learning. We attempt the following two methods that fine-tune backbones with MQ labels.
\emph{One-stage:} The first one is one-stage finetuning. We skip the full Ego4D dataset and directly finetune ir-CSN-152 pre-trained on IG65M or VideoMAE-L pre-trained on K700. 
We denote it as ``K700 $\rightarrow$ MQ''. \emph{Two-stage:} The second method finetunes the backbones on the full Ego4D dataset and then continues finetuning the backbone on the MQ sub-dataset.

%Since we separated the verb and noun annotations from the data level to build two subsets, we need to fine-tune the model with the initial weights separately and obtain two fine-tuned models of verb and noun. The verb and noun models can capture action information and scene information, respectively.

\textbf{Feature Extraction.} \label{feat-extract}
We train the temporal localization methods with offline video features instead of end-to-end training. We extract Spatio-temporal features on videos through a sliding snippet approach at 15 FPS. CSN is adopted to extract a 2048-dimensional feature vector for each snippet. Each snippet contains $s=32$ consecutive frames with snippet interval $\delta=8$. For VideoMAE, we extract a 1024-dimensional feature vector for each snippet that contains $s=16$ consecutive frames with interval $\delta=8$. Further, we separately extract verb and noun features for all backbones and aggregate them to enhance video representations.

\textbf{Localization Head.} We adopt the official baseline VSGN~\cite{vsgn} for this temporal action localization task with MQ. Then we replace its detector with ActionFormer~\cite{actionformer}, further improving the final localization performance.

\begin{table}[t]
\centering
\small
\setlength\tabcolsep{2mm}
\begin{tabular}{c|cc|cc}
\toprule
Feature & \multicolumn{2}{c|}{Validation} & \multicolumn{2}{c}{Test} \\
(K700 $\rightarrow$ MQ) & Recall & mAP & Recall & mAP \\
\midrule 
ir-CSN-152~\cite{csn} & 28.96 & 10.63 & 28.53 & 10.68 \\
VideoMAE-L~\cite{videomae} & 29.93 & 11.95 & 27.71 & 11.44  \\ 
VideoMAE-L~\cite{videomae} (MVF) & 35.19 & 15.11 & 33.50 & 14.26 \\ 
\bottomrule
\end{tabular}
\caption{The Moment Queries performance of VSGN using the features extracted from the one-stage fine-tuned backbone. ``Recall'' and ``mAP'' denote as Recall@1 at tIoU=0.5 and average mAP at tIoU from 0.1 to 0.5. }
\label{table:mq-one-stage-finetune-performance}
\end{table}

\begin{table}[t]
\centering
\small
\setlength\tabcolsep{1.2mm}
\begin{tabular}{c|c|cc|cc}
\toprule
Feature & \multirow{2}{*}{Method} & \multicolumn{2}{c|}{Validation} & \multicolumn{2}{c}{Test} \\
 (VideoMAE-L~\cite{videomae}) &  &Recall & mAP & Recall & mAP \\
\midrule 
K700 $\rightarrow$ Verb $\rightarrow$ MQ & VSGN & 37.82 & 19.35 & 36.38 & 18.04 \\
K700 $\rightarrow$ Verb & AF & 37.24 & 20.69 & 35.58 & 19.31  \\ 
K700 $\rightarrow$ Verb $\rightarrow$ MQ & AF & \textbf{40.36} & \textbf{23.29} & \textbf{41.13} & \textbf{23.59} \\ 
\bottomrule
\end{tabular}
\caption{The Moment Queries performance uses the features extracted from the backbone with or without the second stage of fine-tuning. ``Recall'' and ``mAP'' denote the Recall@1 at tIoU=0.5 and average mAP at tIoU from 0.1 to 0.5. ``AF'' represents ActionFormer.}
\label{table:mq-two-stage-finetune-performance}
\end{table}

\begin{table*}[t]
\centering
\small
\setlength\tabcolsep{1.7mm}
\begin{tabular}{c|c|cccc|cccc}
\toprule
 \multicolumn{1}{c}{\multirow{2}{*}{\#}}&\multicolumn{1}{|c}{\multirow{2}{*}{Method}} & \multicolumn{4}{|c|}{Validation} & \multicolumn{4}{c}{Test} \\
&  & R5@0.3 &R5@0.5 & R1@0.3 & R1@0.5 & R5@0.3 &R5@0.5 & R1@0.3 & R1@0.5 \\
\midrule 
A&EgoVLP~\cite{egovlp} &  18.84 & 13.45 & 10.84 & 6.81 &  16.76 & 11.29 & 10.46 & 6.24\\
B&VideoMAE-Verb~\cite{videomae} + EgoVLP-TE~\cite{egovlp} & 21.73 & 15.07 & 12.32 & 7.43 & 20.32 & 13.29 & 13.03 & 7.87  \\ 
C&VideoMAE-Noun~\cite{videomae} + EgoVLP-TE~\cite{egovlp} & 21.89 & 15.64 & 12.78 & 8.08 & $-$ & $-$ & $-$ & $-$ \\ 
D&(B+C) Pre-fusion & 23.36 & 17.37 & 13.71 & 9.06 & 21.25 & 14.64 & 14.59 & 9.07 \\ 
E&(D+A) Pre-fusion & 24.21 & 17.89 & 14.40 & 9.60 & 21.98 & 15.28 & 15.56 & 9.99 \\ 
F&(D+E) Post-fusion & \textbf{24.78} & \textbf{18.30} & \textbf{15.64} & \textbf{10.17} & \textbf{22.95} & \textbf{16.10} & \textbf{16.45} & \textbf{10.06} \\ 
\bottomrule
\end{tabular}
\caption{The Natural Language Queries performance. ``EgoVLP-TE'' denotes the text encoder of EgoVLP.}
\label{table:nlq-performance}
\end{table*}

\subsection{Experiments}
We perform experiments with different backbones, fine-tuning settings, and localization heads on the validation set and submit inference results on the testing set to EvalAI's test server.

We first evaluate the localization performance with the \emph{one-stage fine-tuning} setting and VSGN.
We conducted some experiments with this method, and the results are listed in Table~\ref{table:mq-one-stage-finetune-performance}. 
Comparing VideoMAE features and CSN features, we find that the former can bring higher mAP, and the latter can achieve higher Recall on the test set. We also explore the spatial-level Multi-View Fusion (identified as ``MVF'' in the table) method to enhance temporal feature representations, improving localization performance but increasing computational overhead.

Finally, we adopt the \emph{two-stage fine-tuning} method as our solution for MQ tasks.
We extract the two types of features (denoted as ``K700 $\rightarrow$ Verb $\rightarrow$ MQ'' and ``K700 $\rightarrow$ Verb'' in the table) with and without the second finetuning stage to explore their effect on localization performance. ActionFormer is used to replace VSGN to unlock the potential of the temporal features. The experiment results on the validation and test set are shown in Table~\ref{table:mq-two-stage-finetune-performance}.

Comparing Table~\ref{table:mq-two-stage-finetune-performance} and Table~\ref{table:mq-one-stage-finetune-performance}, introducing the full Ego4D dataset greatly boosts mAP and Recall. 
As shown in the second and third rows in Table~\ref{table:mq-two-stage-finetune-performance}, the second stage of finetuning transfers the feature representation of the backbone from the lower-level representation of verbs or nouns to the higher-level activity representation of the MQ dataset. It improves the overall localization performance, with +5.55 in Recall and +4.28 in mAP on the test set.

\section{Track 2: Natural Language Queries}

\subsection{Problem Definition}

The Natural Language Queries track serves as the Episodic Memory task to query the moments corresponding to some text. 
For an untrimmed egocentric video, we denote it as $U=\{u_t\}^{l_{v}}_{t=1}$, where $l_v$ indicates the length of the video and  $u_t$ is the $t$-th frame. We denote the temporal annotation of action instances as $\Psi_{\rm g}=\{\varphi_n=(t_{\rm s},t_{\rm e},q_{t})\}^{N_{\rm g}}_{n=1}$ in the video $S_v$ which has  $N_{\rm g}$ instance. $t_{\rm s}$, $t_{\rm e}$, and $q_{t}$ are the start, end boundary, and text query corresponding to the instance $\varphi$, respectively. 
The model generates predicted moment segments that should cover $\Psi_{\rm g}$ and high temporal overlapping.

\subsection{Approach}
Benefiting from the high performance of the VideoMAE feature in the MQ task, we explore its performance in the NLQ task.
We adopt the method mentioned in Sec~\ref{feat-extract} to extract video features. The official baseline VSLNet~\cite{vslnet} is applied to solve this task.

\subsection{Experiments}
Using EgoVLP~\cite{egovlp} as our baseline, we perform several further experiments shown in Table~\ref{table:nlq-performance}. 
Different configurations are identified by capital letters from ``A'' to ``F''. We will introduce these configurations and the improvements they bring.

\textbf{A.} We use EgoVLP~\cite{egovlp} as our baseline.

\textbf{B \& C.} We first replace the video feature of EgoVLP with our verb and noun features, respectively, and preserve the text encoder of EgoVLP. Comparing EgoVLP, our video features improve by about +2.0 R1@0.3 and R1@0.5. Meanwhile, the noun feature is better than the verb feature, which suggests some domain gaps between the verb feature and the noun feature.

\textbf{D.} Based on the previous experiment, we assume that the verb and noun features pay more attention to the motion and scene information in videos, respectively.
We fuse the verb and noun features before feeding them into the model. It brings at least a +1.0 R1 improvement, proving that verb and noun features are complementary.

\textbf{E.} We further tried to fuse the video features of EgoVLP and gained about +0.9 R1@0.3 and R1@0.5 improvement on the test set. Since EgoVLP is a multi-modal pre-training method, it may be complementary to full-supervised pre-training.

\textbf{F.} Finally, we post-fuse the predictions of D and E to improve the overall NLQ performance.

\section{Track 3: Future Hand Prediction}

\begin{table*}[t]
\centering
\small
\setlength{\tabcolsep}{2.4mm}
\begin{tabular}{c|cc|cc|cc|cc}
\toprule
  & \multicolumn{4}{c|}{Validation} & \multicolumn{4}{c}{Test}  \\ 
% \midrule
 Method & \multicolumn{2}{c|}{Left Hand}  & \multicolumn{2}{c|}{Right Hand}  & \multicolumn{2}{c|}{Left Hand}  & \multicolumn{2}{c}{Right Hand }   \\
 & M.Disp.$\downarrow$ & C.Disp.$\downarrow$ & M.Disp.$\downarrow$ & C.Disp.$\downarrow$  & M.Disp.$\downarrow$ & C.Disp.$\downarrow$ & M.Disp.$\downarrow$ & C.Disp.$\downarrow$  \\
\midrule 
I3D~\cite{i3d} (224,16,30) & 54.11 & 57.29 & 54.73 &57.94 & 52.98 & 56.37 & 53.68 & 56.17 \\ 
VideoMAE-L~\cite{videomae} (224,16,1) & 66.45 & 68.23 & 67.32 & 68.92 & $-$ & $-$ & $-$ & $-$ \\ 
UniFormer-B~\cite{uniformer} (320,4,1) & 46.65 & 54.58 & 48.30 & 55.10 & 45.76 & 54.95 & 47.93 & 55.11 \\ 
UniFormer-B~\cite{uniformer} (320,4,30) & 44.90 & 54.16 & 46.70 & 54.66 & 44.69 & 53.47 & 47.00 & 53.49 \\ 
UniFormer-B~\cite{uniformer} (320,8,30) & \textbf{43.25} & \textbf{52.78} & \textbf{45.29} & \textbf{52.65} & \textbf{43.85} & \textbf{53.33} & \textbf{46.25} & \textbf{53.37} \\ 
% UniFormer-B (320,16,3) & 45.92 & 54.31 & 48.53 & 55.32 & - & - & - & - \\ 
\bottomrule

\end{tabular}
\caption{The Future Hand Prediction performance of 
 UniFormer-B. In the first column, the testing settings denote as (S, T, V)} for the size of frames, the number of frames, and the number of temporal views.
\label{table:hands-results}
\end{table*}

\subsection{Problem Definition}
The Future Hand Prediction track serves as the forecasting task to forecast the spatial location of future hands. 
Specifically, we denote the contact frame as $x_c$, the pre-condition frame as $x_p$, and the three frames preceding the pre-condition frame by $0.5s$, $1s$, and $1.5s$ as $x_{p1}$ , $x_{p2}$ , $x_{p3}$ , respectively. Formally, given an input egocentric video $1.5s$ before the pre-condition time step (denoted as $x=\{x_{p_{3}-t_{o}-1},...,x_{p_3-1}\}$, with $t_o$ referred as observation time), this task seeks to predict the positions of both hands $(h^l_i, h^r_i )$ in the future key frames, where $i \in \{c, p, p_1, p_2, p_3\}$.

\subsection{Approach}
We follow the baseline method, predicting 20 categories for short-term (1.5 seconds) historical information. The 20 categories indicate the future's absolute spatial coordinates of five pairs of hands. We use L1 loss to regress these coordinates.

Inspired by the locality introduced by the convolution module in object detection benefits the regression boundary, we adopt UniFormer~\cite{uniformer} as our backbone, which contains depth-wise convolution modules to preserve the local spatial information explicitly. This historical spatial local information is a major source for regressing the coordinates in the future.

\subsection{Experiments}

We attempt to use VideoMAE-L and UniFormer-B to regress the spatial location of the future hands. The results are shown in Table~\ref{table:hands-results}. 
We first perform experiments using VideoMAE-L that only contains global self-attention modules to ablate the effect of convolution modules. 
When training VideoMAE-L, the network is difficult to converge, and the final result is weaker than I3D~\cite{i3d}.

We use UniFormer-B as the backbone of our final solution. 
The forecasting results have been much better when inputting 4 frames with 320$\times$320 resolution than I3D. 
It may be because of higher spatial resolution or a stronger convolution hybrid backbone. Furthermore, we increase the number of input frames and fuse multi-view predictions to enhance the regression performance.

\section{Track 4: State Change Object Detection}
\subsection{Problem Definition}

State Change Object Detection (SCOD) is the task of detecting the object undergoing a state change from the given egocentric video clips.
Specifically, each given video consists of three temporal frames, \ie~precondition (PRE), point-of-no-return (PNR), and post-condition (POST), and the goal is to predict the 2D bounding boxes of the state change object in the PNR frame.

\subsection{Approach}
For this task, the officials provide several single-frame baselines covering a wide range of detection frameworks, including CenterNet~\cite{centernet}, Faster R-CNN~\cite{faster-rcnn}, and DETR~\cite{detr}.
These methods adopt ResNet~\cite{he2016deep} and DLA~\cite{dla} as the backbones and achieve decent results. 
Nevertheless, we argue that the above methods still have some room for improvement due to the sub-optimal choice of the backbone network and detection head.

In this technical report, we develop a stronger detector for the SCOD task.
Specifically, we follow the official baseline to build a single-frame detector and perform 2D object detection on the PNR frame.
Our method consists of two key components: (1) an image backbone (\eg, UniFormer-L~\cite{uniformer} or Swin-L~\cite{swin}), and (2) a query-based detection head DINO~\cite{dino}.

On this basis, we further explore the transfer learning from general object detection to egocentric images, including three different pre-training tasks of ImageNet classification~\cite{deng2009imagenet}, COCO detection~\cite{cocodet}, and Objects365 detection~\cite{object365} (\ie, a larger detection dataset).
The experiments indicate that our method achieves a promising improvement over the official baselines, and the pre-training on general object detection can derive significant benefits for the SCOD task. 
We hope this method can serve as a strong baseline for egocentric object detection.

\begin{table*}[t]
\centering
\small
\setlength{\tabcolsep}{2.0mm}
\begin{tabular}{ccc|ccc|ccc}
\toprule
\multicolumn{1}{c}{\multirow{2}{*}{Method}} & \multicolumn{1}{c}{\multirow{2}{*}{Detector}} & \multicolumn{1}{c}{\multirow{2}{*}{Pre-training Dataset}} & \multicolumn{3}{|c|}{Validation Set}  & \multicolumn{3}{c}{Test Set}  \\
\multicolumn{1}{c}{}& \multicolumn{1}{c}{}                          &       & AP & AP50 & AP75 & AP & AP50 & AP75 \\
\midrule 
ResNet-101~\cite{he2016deep} & Faster R-CNN~\cite{faster-rcnn} &ImageNet-1K~\cite{deng2009imagenet} &13.40 & 25.60 & 12.50 &13.35 & 25.52 & 12.38  \\ 
 ResNet-50~\cite{he2016deep} &DETR~\cite{detr} & ImageNet-1K~\cite{deng2009imagenet} &15.50 & 32.80 & 13.00 &15.38 &32.51 &12.87  \\ 
 DLA-34~\cite{dla} & CenterNet~\cite{centernet} & ImageNet-1K~\cite{deng2009imagenet} &6.40 & 11.70 & 6.10 &6.32 &11.62 & 6.08 \\ 
\midrule
UniFormer-L~\cite{swin} & DINO~\cite{dino} & ImageNet-1K~\cite{deng2009imagenet} & 24.80 & 44.20 & 24.00 & $-$ & $-$ & $-$ \\ 
Swin-L~\cite{swin} & DINO~\cite{dino} & ImageNet-22K~\cite{deng2009imagenet} & 28.00 & 48.70 & 27.20  & $-$ & $-$ & $-$ \\ 
Swin-L~\cite{swin} & DINO~\cite{dino} & ImageNet-22K~\cite{deng2009imagenet} + COCO~\cite{cocodet} & 32.20 & 51.30 & 33.10 & $-$ & $-$ & $-$   \\
Swin-L~\cite{swin} & DINO~\cite{dino} & ImageNet-22K~\cite{deng2009imagenet} + Objects365~\cite{object365} & \textbf{36.40} & \textbf{56.50} & \textbf{37.60}  & \textbf{37.19} & \textbf{55.97} & \textbf{38.44}  \\

\bottomrule

\end{tabular}
\caption{The detection performance of official baselines and our method on SCOD validation and test set.}
\label{table:scod-performance}
\end{table*}

\subsection{Experiments}

Our detection experiments are based on the SCOD dataset and the MMDetection~\cite{mmdetection} codebase.
We adopt UniFormer-L~\cite{uniformer} pre-trained on ImageNet-1K, or Swin-L~\cite{swin} pre-trained on ImageNet-22K as the backbone.
In addition, we employ DINO~\cite{dino} as the detection head, in which the numbers of content and denoising queries are fixed to 900 and 1000, respectively. 

Firstly, we tried to train the SCOD dataset without extra detection datasets. 
Further, we study the transfer learning from two general object detection datasets (\ie, COCO~\cite{cocodet} and Objects365~\cite{object365}) to the SCOD dataset. The pre-training schedules for these two datasets are 12 epochs and 26 epochs, respectively.
During SCOD fine-tuning, the shorter side of the input image is resized between 800 and 1600, while the longer side is at most 2000. 
All models are trained with AdamW optimizer (batch size of 16, initial learning rate of 1$\times$10$^{-4}$, and weight decay of 0.0001) for 12 epochs.

As shown in Table~\ref{table:scod-performance}, when using only ImageNet-22K~\cite{deng2009imagenet} pre-training, our method yields an impressive score of 28.0 AP on the SCOD validation set, outperforming previous official baselines by at least +12.5 AP. 
We can also see that pre-training on general object detection can greatly benefit the SCOD task.
For instance, COCO~\cite{cocodet} pre-training promotes the detection performance to 32.2 AP, and Objects365~\cite{object365} pre-training further achieves a big jump to 36.4 AP.
We submitted the best result to the test server, achieving 37.2 AP on the test set and ranking 1$^{\text{st}}$ on the leaderboard.

%------------------------------------------------------------------------
% \subsection{Conclusion}
% This report builds a powerful detector with the widely-used backbone UniFormer-L/Swin-L and the state-of-the-art detection head DINO. Further, we explore the transfer learning from general object detection to egocentric images. 
% Our model achieves a big jump to 37.2 AP on the test set and ranks 1$^{\text{st}}$ on the leaderboard of Ego4D SCOD Challenge 2022.
% We hope this simple yet powerful method can serve as a strong baseline for egocentric object detection.

% \textbf{Limitation.} Our method only uses PNR frames, which may not fully satisfy the original motivation for this task and may depend on pre-training on large-scale object detection datasets. Meanwhile, our method uses a relatively large backbone and a complex detection head, which may not meet the real-time requirements.

%\input{content/track-longterm}
\section{Track 5: Short-term Object Interaction Anticipation}
\subsection{Problem Definition}
\label{sec:intro}
The Short-term hand object prediction task aims to predict the next human-object interaction happening after a given timestamp. For a given video, we need to detect the spatial location of active objects and perform noun classification. These active objects will be considered to interact with people at the $\Delta$ time in the future. The action category of the interaction also needs to be identified, and the value of the $\Delta$ needs to be predicted.

\begin{table*}[t]
\centering
\small
\setlength\tabcolsep{1.87mm}
\begin{tabular}{c|c|cccc|cccc}
\toprule
\multicolumn{1}{c}{\multirow{2}{*}{\#}}&\multicolumn{1}{|c}{\multirow{2}{*}{Method}} & \multicolumn{4}{|c|}{Validation} & \multicolumn{4}{|c}{Test} \\
&  & Noun &Noun+Verb &Noun+TTC & Overall& Noun &Noun+Verb &Noun+TTC & Overall \\
\midrule 
A&Baseline~\cite{ego4d} &17.55&5.16&5.19&1.98&20.45 &6.63 &5.93 &2.20 \\
B&VideoMAE-L~\cite{videomae} &17.55&5.37&5.21&2.06& 20.45 & 7.84 & 5.74 & 2.38   \\ 
C&VideoMAE-L~\cite{videomae} + Box-embed &17.55&6.30&5.83&2.43& 20.45 & 7.64 & 6.85 & 2.88  \\ 
D&C+new top3box &18.73&8.5&7.55&3.87& 20.46 & 7.39 & 7.18 & 3.00  \\ 
E&C+new box+fusion &20.02&7.34&6.37&2.74& 24.53 &9.09 & 7.59 & 3.36  \\ 
F&C+new top10box+fusion&19.45&8.00&6.97&3.25& \textbf{24.60} & \textbf{9.18} & \textbf{7.64} & \textbf{3.40} \\ 
\bottomrule
\end{tabular}
\caption{The Short-term object interaction anticipation performance.}
\label{table:short-term}
\end{table*}

\subsection{Approach}
In this task, the baseline method uses a two-stage approach, which is commonly used in ST-AL tasks. Given a series of video keyframes~$(k_1,k_2,\cdots ,k_n)$, these keyframes can be extracted from the same video or different videos. 
In the first stage, the object detector is trained first, and the detector generates a series of localization boxes from the keyframes of the video.
We then predict a noun category for each bounding box. 
In the second stage, for each keyframe, we sample a clip before it to get a series of frames and feed them into the backbone to extract features. 
In the final classification stage, we use RoIAlign to capture areas of interest from the extracted features, classify actions, and regress contact time based on these features. 

In the baseline method, Faster R-CNN~\cite{faster-rcnn} is used for detection in the first stage. In the second stage, the baseline uses SlowFast~\cite{slowfast} to perform feature extraction on the input video and classify it at the end. 
As we mentioned before, this task needs to predict actions and time to contact, both of which are related to the interaction between people and objects. 
However, the receptive fields of the convolutional neural network are limited, and capturing people's interactions with things is difficult. 
In addition to the limitation in capturing the interaction relationship, the baseline method directly uses RoI features to predict time to contact, which is an indication of the time of human-object interaction and is related to the location where the human and the object are located. It is not reasonable to use RoI features only to predict the time.

Based on these considerations, we have made the following improvements:

(1) We use the ViT~\cite{vit} with space-time attention pre-trained by VideoMAE~\cite{videomae} to capture the interaction between people and objects. It exploits self-attention with long-range modeling capability to characterize such an interaction, which benefits action recognition.

(2) To better combine the time to contact with the position of the box, we perform a positional encoding operation on the boxes and fuse the positional information of the box with the RoI feature to predict the time to contact.

(3) Although the bounding boxes provided by Ego4D officials already achieved decent performance, we used our own trained detectors in the first stage to further improve the prediction quality and then conducted some post-processing to reduce redundant boxes.

\subsubsection{Detector Training}
We used the open-sourced detection framework MMDetection\cite{mmdetection} to train the current state-of-the-art detector DINO \cite{dino}. DINO \cite{dino} is an improved object detection algorithm based on DETR\cite{detr}. It uses a transformer network to generate a fixed number of bounding boxes per keyframe. In this task, we generate 900 boxes for each keyframe and then perform the non-maximum suppression (NMS) post-processing to remove redundant predictions.

\subsubsection{Video Backbone}
We used VideoMAE~\cite{videomae} as the backbone, which is a complete transformer network. 
VideoMAE is pre-trained on the K700 dataset and then trained for basic classification on the Ego4D~\cite{ego4d} dataset. The Transformer-based network is chosen because of its explicit attention mechanism and competitive performance in various discriminative tasks. 
It is effective for modeling the long-range relationship between people and objects and is conducive to the recognition of human actions.
After the pre-training of VideoMAE, we finetune the encoder to extract features from the input video in the whole pipeline. 

Although we already have well-performing boxes in the first stage, these boxes are still quite different from the ground truth. 
In the original configuration of the baseline method, only the boxes predicted in the first stage are used for training.
Since the boxes are not accurate enough, it is highly likely to bring inevitable errors to the second stage, degrading its training. 
The Spatio-Temporal localization task meets a similar problem as well. 
The common practice is using the ground-truth boxes for training and the boxes predicted by the detector for validation and testing. 
However, due to the small number of ground truth boxes in each frame in the training set, the amount of training data will be greatly reduced. To improve the generalization ability of the model, we finally adopt the method of taking both ground truth boxes and predicted boxes as input. 
We find that it produces better results than using either ground truth or predicted boxes alone.

We initially followed the baseline method to directly perform the RoIAlign operation on the input box to obtain the RoI feature, and predict the verb category and time to contact for each RoI feature. The verb prediction performance is better than the baseline method, but for time to contact, using the RoI feature directly for prediction has a trivial improvement. After analysis, we found that since the convolutional neural network itself has position prior information, which is very beneficial for the prediction of time to contact depending heavily on the position, so the RoI features extracted by the SlowFast~\cite{slowfast} framework are more suitable. 

To improve the prediction performance of time to contact, we first tried to use the boxes directly for prediction. 
Considering that there are clip and flip operations in data augmentation, they will change the position of boxes, so we use the raw boxes in the prediction.
We normalize and input them to the MLP~\cite{transformer}, and the prediction results are obtained. However, after testing, we observed that it is still not as good as the baseline method.
This shows that it is not possible to use only the position information. 
We consider fusing the position information with the extracted features. 
Using the position encoding operation of the transformer, we first encode the position of the box by cosine position encoding and encode the position of the box. 
Then we add it to the corresponding RoI feature to complete the fusion. Such a fusion is reasonable, which is validated in Section~\ref{experiments}.

\subsubsection{Box Processing and Result Fusion}
Since the generated boxes in the first stage are redundant, we need to further adjust the box number before fine-tuning the classification model. 
Considering that the evaluation metric is the boxes with the top-5 scores, we filtered out the top 5 boxes using the noun classification score. 
This operation greatly reduces the number of boxes and speeds up the inference. 

Using the filtered boxes for prediction, the results on the validation set have all surpassed the baseline. To fully use existing boxes and prediction results, we fuse the prediction results using our boxes and the official predictions. The specific fusion method is to directly splice the prediction results of the corresponding keyframes in the two result files. 
This will introduce both boxes that are relatively similar to our box, as well as some less effective boxes. 
The score of the box with poor effects is usually relatively low, and it will be automatically filtered out in the evaluation program, and similar boxes are redundant. Here, NMS~\cite{nms} is used again, and we set a higher IoU threshold to remove redundant boxes.

\subsection{Experiments}
\label{experiments}

\textbf{Implementation Details.}
We use VideoMAE-L~\cite{videomae} as our backbone. The model is trained for 10 epochs using the AdamW optimizer and cosine learning rate schedule. In the fine-tuning stage, we use 8 GPUs for training, with a total batch size of 64, weight decay of 0.05, and a learning rate of $2.5\times 10^{-4}$ for 30 epochs.
% \begin{table}
%   \centering
%   \small
%   \setlength\tabcolsep{5mm}{
%   \begin{tabular}{@{}lc@{}}
%     \toprule
%     Setting & Value \\
%     \midrule
%     Model & VideoMAE-Large \\
%     Resolution & 224 \\
%     Frames & 16 \\
%     Weight decay & 0.05\\
%     \bottomrule
%   \end{tabular}
%   }
%   \caption{Other settings of VideoMAE-L.}
%   \label{tab:vit_l}
% \end{table}

\textbf{Results.}
We first reproduced the baseline according to the configurations and codes provided by the baseline. 
% In the end, we did not reproduce the baseline results in the leaderboard, and there were some gaps.
% \begin{table}[h]
%   \centering
%   \small
%   \setlength\tabcolsep{5mm}
%   \begin{tabular}{@{}lc@{}}
%     \toprule
%     Metrics & Value \\
%     \midrule
%     Noun mAP & 20.45 \\
%     Noun+Verb mAP & 6.63 \\
%     Noun+TTC mAP & 5.93 \\
%     Overall & 2.20\\
%     \bottomrule
%   \end{tabular}
%   \caption{Results of our replicated baseline. ``Overall" means Box+Noun+Verb+TTC mAP.}
%   \label{tab:baseline_ours}
% \end{table}
To improve the performance, we migrated the task to the codebase of VideoMAE~\cite{videomae}. The models used in the experiments described below are all trained from this codebase.
Through experiments, we found that using the transformer network and introducing the positional encoding of the boxes for training is effective for this task.

As shown in Table~\ref{table:short-term}, when filtering the boxes, we not only took the top 5 but also tried to keep the top 3 and top 10 boxes, respectively.
Although keeping the top-3 boxes on the validation set gave the best results, it performed mediocre on the test set. 
It shows that there is a certain gap between the data of the test set and the validation set, and some of our additional operations overfit the validation set.
For the result fusion, we found that when the results of the top 10 boxes are selected for fusion, it produces a better effect, but it does not work when the top 3 boxes are used for fusion.

% \subsection{Conclusion}
% Compared with the previous Spatio-Temporal localization, the biggest challenge of this task is the prediction of time to contact. It is effective to introduce position coding to the box and integrate it with the RoI feature, but the effect is also very limited. We will continue to explore this problem in the future. In addition, since Ego4D data are all in first-person view, they differ from the previous K700 and other datasets. This difference makes common transfer learning difficult. Besides, verb category prediction is closely related to upstream pre-training. It may be why our verb category prediction results are not good enough, which is also an explorable issue.

%\input{content/experiments}

% !TeX spellcheck = en_US
%!TEX root=../main.tex
% \clearpage
\section{Concluding Remarks}
We have presented our solutions to five tracks in the Ego4D ECCV2022 Challenge. We find a strong video backbone can give an advantage to task performance, and video backbones with various structures complementary to each other in these tasks. The typical video dataset has a distinct domain gap from the egocentric one, leading to degraded performance. We close this gap by finetuning the pre-trained video backbones on the Ego4D dataset. 
\paragraph{Limitation} Our feature aggregation method is a bit naive as we directly concatenate features from different backbones. The good craft of feature alignment or other dynamic fusion modules could benefit the egocentric video representation. Besides, adapting the employed backbone with task heads to each track is tedious. How to achieve satisfying task performance while minimizing downstream tweaking remains open.

%\input{content/acknowledge}

%------------------------------------------------------------------------

%%%%%%%%% REFERENCES
{\small
\bibliographystyle{ieee_fullname}
\bibliography{egbib}
}

\end{document}